\documentclass[journal]{IEEEtran}
\usepackage{amsmath,amsfonts}
\usepackage{algorithmic}
\usepackage{algorithm}
\usepackage{array}
\usepackage[caption=false,font=normalsize,labelfont=sf,textfont=sf]{subfig}
\usepackage{textcomp}
\usepackage{stfloats}
\usepackage{url}
\usepackage{verbatim}
\usepackage{graphicx}
\usepackage{booktabs}
\usepackage{cite}
\usepackage{amssymb}
\usepackage{makecell}
\usepackage{multirow} 
\usepackage{color} 
\usepackage{xcolor}
\usepackage[utf8]{inputenc}

\begin{document}

\title{Leveraging Static Relationships for Intra-Type and Inter-Type Message Passing in Video Question Answering}

\author{{Lili Liang, Guanglu Sun}
	
	\thanks{Manuscript received March 24, 2025;}
	\thanks{This work was supported in part by the Key Research and Development Project of Heilongjiang Province with 2022ZX01A34, in part by the Natural Science Foundation of China with No.60903083. (\textit{Corresponding authors: Guanglu Sun}).}
	\thanks{Guanglu Sun and Lili Liang are affiliated with the School of Computer Science and Technology, Harbin University of Science and Technology, Harbin 150080, China; and the Heilongjiang Provincial Key Laboratory of Intelligent Information Processing and Application, Harbin 150080, China (e-mail: sunguanglu@hrbust.edu.cn, lianglili0519@163.com).}
}

\markboth{}%
{Shell \MakeLowercase{\textit{et al.}}: A Sample Article Using IEEEtran.cls for IEEE Journals}


\maketitle

\begin{abstract}
Video Question Answering (VideoQA) is an important research direction in the field of artificial intelligence, enabling machines to understand video content and perform reasoning and answering based on natural language questions. Although methods based on static relationship reasoning have made certain progress, there are still deficiencies in the accuracy of static relationship recognition and representation, and they have not fully utilized the static relationship information in videos for in-depth reasoning and analysis. Therefore, this paper proposes a reasoning method for intra-type and inter-type message passing based on static relationships. This method constructs a dual graph for intra-type message passing reasoning and builds a heterogeneous graph based on static relationships for inter-type message passing reasoning. The intra-type message passing reasoning model captures the neighborhood information of targets and relationships related to the question in the dual graph, updating the dual graph to obtain intra-type clues for answering the question. The inter-type message passing reasoning model captures the neighborhood information of targets and relationships from different categories related to the question in the heterogeneous graph, updating the heterogeneous graph to obtain inter-type clues for answering the question. Finally, the answers are inferred by combining the intra-type and inter-type clues based on static relationships. Experimental results on the ANetQA and Next-QA datasets demonstrate the effectiveness of this method.
\end{abstract}

\begin{IEEEkeywords}
Video question answering, static relationships, message passing, intra-type/inter-type Reasoning, scene graph generation
\end{IEEEkeywords}
\section{Introduction}

\IEEEPARstart{I}{n} the realm of video question answering (VQA), accurately modeling and reasoning about the relationships between objects within a video is a critical task\cite{pise2021relational}. these methods that rely on object similarity and spatiotemporal information to construct graph models often struggle to precisely identify static relationships within the video, especially in the absence of explicit labels. This limitation hinders the model's ability to reason effectively about the video content.

\label{sec:intro}
\begin{figure}[htbp]
	\includegraphics[width=90mm]{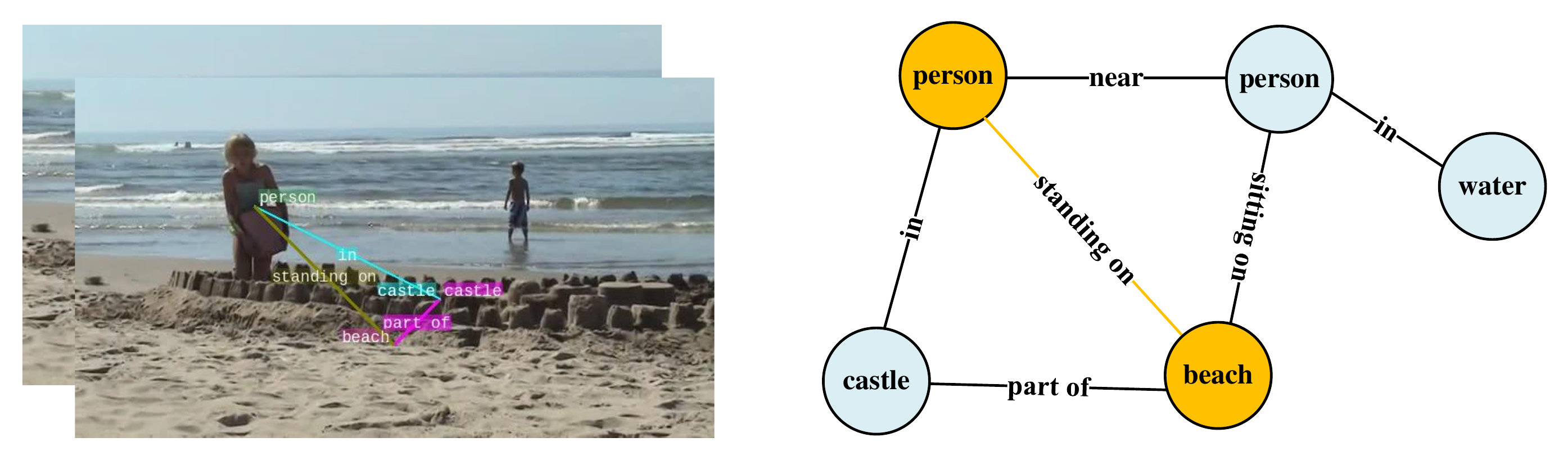}
	\caption{ Example of static relationships in video: the relationship between “person” and “beach” is “standing on” .}
	\label{demo}
\end{figure}

Taking HGA\cite{jiang2020reasoning} and HOSTR\cite{dang2021hierarchical} as examples, although they employ Graph Convolutional Networks (GCNs) to model and reason about the semantics or spatiotemporal positions of video objects, they primarily rely on question labels to guide the reasoning process, which limits their ability to capture static relationships. Therefore, the first research question is how to effectively model the static relationships in videos to improve the accuracy of object and relationship recognition.

Methods for graph reasoning based on object and relationship labels have made certain progress in academia. For example, the LGCN\cite{hu2019language} model utilizes the changes of question-focused objects in the message passing process to obtain clues related to the question. However, LGCN only updates node information during message passing, ignoring the update of edge information in the graph. As a result, the obtained contextual information only includes node context, but not relationship context. D-Transformer\cite{cherian20222} constructs a complex graph model containing 2.5D static features and 1D dynamic features, and implements an efficient reasoning process on the graph model using Transformer. SHG-VQA\cite{urooj2023learning} builds a scenario hypergraph based on videos, which includes objects and relationships between objects, and performs reasoning on the hypergraph using cross-modal Transformer. However, these methods have limitations in handling video content. They treat all instances of objects and relationships as the same category, focusing mainly on intra-type interaction information while neglecting inter-type interactions between object and relationship instances. Therefore, the second research question is how to effectively utilize static relationships in videos for both intra-type and inter-type reasoning to better adapt to the reality and complexity of video content.

To tackle these questions, we propose a novel framework called Type-Aware Message Passing (TAMP). This framework introduces a dual graph structure, consisting of a dual graph and a heterogeneous graph, to facilitate both intra-type and inter-type message passing. The dual graph captures the intra-type relationships by passing messages within the object and relationship classes, while the heterogeneous graph captures the inter-type relationships by passing messages between objects and relationships. This comprehensive approach allows the model to effectively reason about the video content by considering both the intra-type and inter-type static relationships.

By integrating both intra-type and inter-type static relationships, our TAMP framework provides a comprehensive reasoning approach that significantly improves the accuracy and robustness of video question answering models. The main contributions of this chapter can be summarized as follows:

\begin{itemize}[\leftmargin=0.85cm]
	\item[$\bullet$] We introduce a video question answering framework based on intra-type and inter-type message passing reasoning using static relationships. This framework constructs a dual graph for intra-type message passing and a heterogeneous graph for inter-type message passing, effectively modeling and reasoning about the static relationships related to the question in the video.	
	\item[$\bullet$] We propose an intra-type message passing reasoning model that performs message passing within the object and relationship classes. By capturing the neighborhood information of objects and relationships related to the question in the dual graph, we update the dual graph to obtain intra-type clues for answering the question.
	\item[$\bullet$]  We propose an inter-type message passing reasoning model that performs message passing between objects and relationships in both directions. By capturing the neighborhood information of different categories of objects and relationships related to the question in the heterogeneous graph, we update the heterogeneous graph to obtain inter-type clues for answering the question.
	\end{itemize}

\section{Related Work}
\label{sec:formatting}
\subsection{Video Question Answering }
In recent years, the predominant approach in VideoQA has been to deduce answers by aligning visual and textual data. Common methods utilize cross-attention mechanisms \cite{xu2017video} to facilitate this alignment across different modalities, and they perform reasoning via either multi-hop attention \cite{zhao2018multi, le2019learning} or stacked self-attention layers \cite{li2019beyond}.

Recently, pre-trained Transformer models have enhanced cognition and reasoning in VideoQA. Various approaches, such as BERT \cite{yang2020bert} and MMFT \cite{urooj2020mmft}, integrate multimodal tokens and process them through self-attention layers, thereby improving cross-modal cognition via attention mechanisms. However, these methods often neglect the temporal occurrences essential for temporal reasoning. To rectify this, models like All in One \cite{wang2023all}, PMT \cite{peng2023efficient}, and RTransformer \cite{zhang2020action} focus on capturing and utilizing temporal information to strengthen model comprehension and reasoning abilities. For more sophisticated reasoning, MIST \cite{gao2023mist} and VGT \cite{xiao2022video} employ strategies like multi-step spatio-temporal reasoning and graph-based reasoning to reveal the intrinsic structure within videos. HSTT \cite{bai2024event} explicitly applies the structure prior of the event graph during both the input and encoding stages to enhance reasoning capabilities. However, current models typically depend on neural architectures for multi-step reasoning, which limits their ability to provide explanations.

\subsection{Scene Graph Generation}
Although the SGG task has made progress, the long-tail distribution of relations leads models to favor the general relations. This limits the application of the SGG task in other fields. To counter this, recent studies \cite{zhao2021semantically,zheng2023dual,li2023label} focus on unbiased scene graph generation. BGNN \cite{li2021bipartite} utilizes the re-sampling strategy to enrich the samples. TGDA\cite{zang2024template} generates relation templates based on knowledge distillation to provide supplementary training data. The data enhancement techniques designed by these methods, while beneficial for the entire dataset, may not address the issue of long-tail distribution. Therefore, some method \cite{liu2023importance} introduces the re-weighting strategy to adjust weights for each relation, making models favor the tail classes. However, these studies neglect to maintain competitive performance on head classes.

In this work, TAMP improves the predictive performance on both head and tail classes by modeling and capturing the interactions for different relations, respectively.

\subsection{Massage Passing}

To improve the prediction of objects and relations in SGG, some works \cite{lin2020gps,li2021bipartite} combine the directional information of relations to aggregate direction-aware context via the graph neural network. Some methods\cite{yoon2023unbiased,yang2024adaptive} use the attention mechanism to adaptively adjust the aggregation weight and aggregate context from neighboring objects to refine the features of objects and relations. Some studies\cite{zhong2021learning,chiou2021recovering} pass messages between visual features and textual embedding by Transformer or LSTM. However, these methods only focus on the context between objects and relations (that is, inter-type message passing between objects and relations) and ignore the semantic context among different objects with the same relation and among different relations with the same object (that is, intra-type message passing of objects and intra-type message passing of relations). This limits these models to fully understanding and representing complex interactions within a scene, thereby reducing the accuracy in relation prediction. 
Unlike previous methods, TAMP captures the intra-type context and the inter-type context, thereby enhancing the understanding of the scene and improving the accurate prediction of relations. 

\section{METHOD}
\begin{figure*}[htbp]
	\centering
	\setlength{\belowcaptionskip}{-0.3cm}
	\includegraphics[width=170mm,height=85mm]{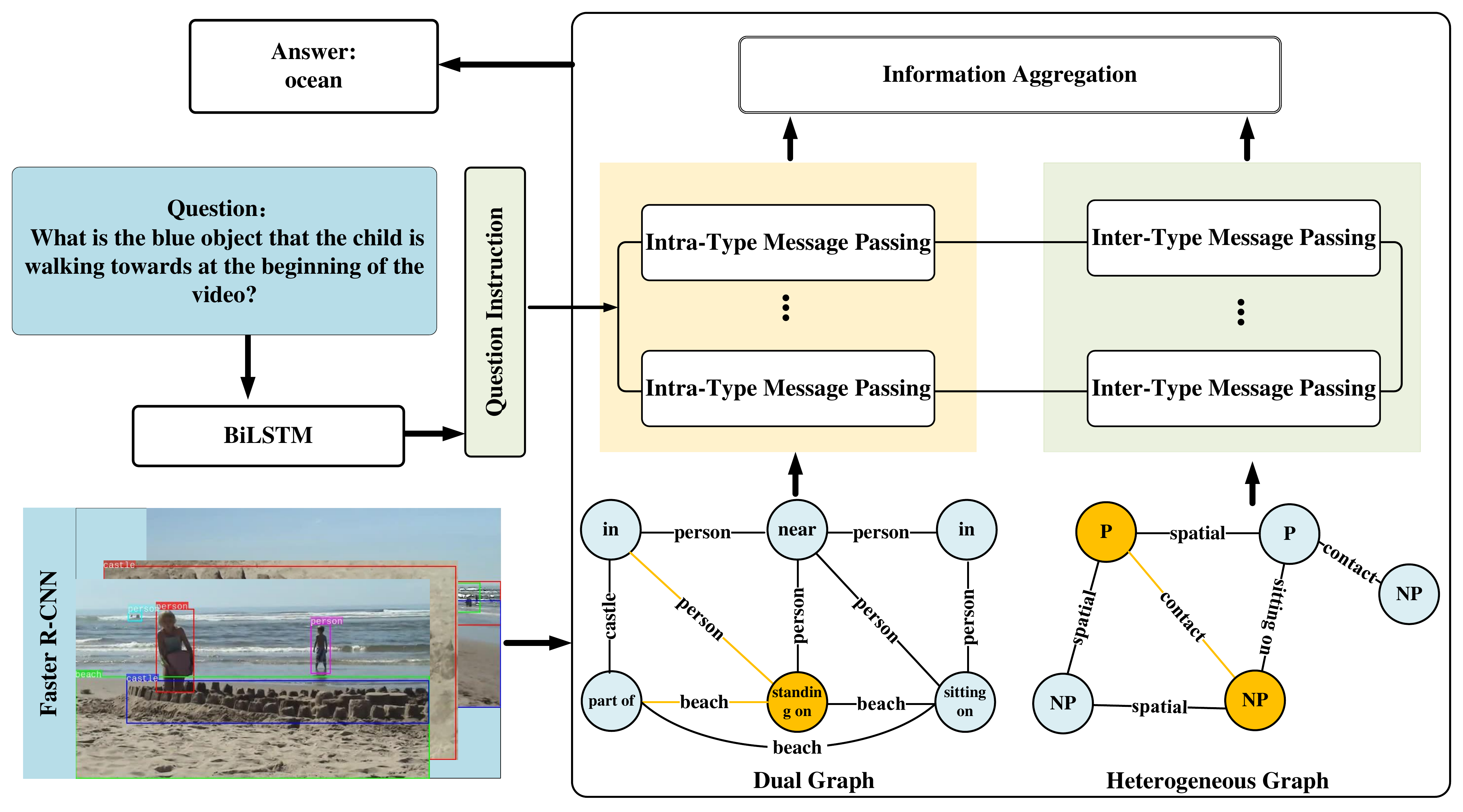}
	\caption{The framework of intra- and inter-type message passing reasoning based on static relationships.}
	\label{overview}
\end{figure*}

\subsection{Question Instruction Generation}
To utilize question guidance for message passing, we create a question instruction for each transmission. Specifically, given the question representation $Q = \{q_1, q_2, ..., q_s\}$, where $q_s$ represents the word embedding of the $s$-th word in the question, and $S$ denotes the length of the question. We use a BiLSTM network to obtain a continuous question representation $q = \{h_1, h_2, ..., h_s\}$, with the specific calculation method as follows:

\begin{equation}
	(h_1, h_2, ..., h_s) = \text{BiLSTM}(q_1, q_2, ..., q_s)
\end{equation}

\begin{equation}
	q = [h_1; h_s]
\end{equation}

At each time step, the hidden states from two directions are stacked to obtain the hidden representation $h_s = [h_s; \overline{h_s}]$ at time $s$, where $[\cdot; \cdot]$ denotes concatenation. During the $l$-th iteration of message passing, the text attention $\alpha_{l,s}$ is calculated for each word in the question, and the question instruction $c_l$ is obtained from the text attention.

\begin{equation}
	\alpha_{l,s} = \text{softmax}\left(\mathbf{W}_s \left(h_s \odot \left(\mathbf{W}_q^{(l)} \text{ReLU}(\mathbf{W}_q q)\right)\right)\right)
\end{equation}

\begin{equation}
	c_l = \sum_{i=1}^L \alpha_{l,z} \cdot h_z
\end{equation}

Here, $\odot$ denotes element-wise multiplication, $\mathbf{W}_s$ and $\mathbf{W}_q$ are shared weight matrices, and $\mathbf{W}_q^{(l)}$ is the weight matrix learned during the $l$-th iteration. $c_l$ can be considered as the question instruction provided during the $l$-th iteration of message passing.

\subsection{Dual Graph Construction Based on Static Relationships}

This section builds a dual graph based on static relationships. It not only maintains the structure of the original graph but also reverses the roles of nodes and edges, allowing the model to learn in a relationship-centered way across different video scenes, thereby capturing and learning the rich contextual information between objects.

The dual graph is defined as $G_{\alpha} = \{V, E\}$, where $V$ represents the set of all relationships, and $E$ represents the set of all nodes. This section uses the Faster R-CNN model to extract target information from the video and construct the initial graph $G$. Specifically, for the detected target $i$, obtain its visual features $v_i$, bounding box features $b_i$, and class label features $c_i$. In the initial graph $G$, the node features of the target are calculated by the following formula:

\begin{equation}
	x_i = \mathbf{W}_c [\mathbf{W}_v v_i; \mathbf{W}_b b_i; c_i]
\end{equation}

where $\mathbf{W}_v$, $\mathbf{W}_b$, and $\mathbf{W}_c$ are learnable linear transformation matrices in the model. For the relationship between target $i$ and target $j$, denoted as $r_{i \rightarrow j}$, its relationship features $x_{i \rightarrow j}$ in the initial graph $G$ can be defined as:

\begin{equation}
	x_{i \rightarrow j} = \mathbf{W}_r [x_i; x_j; b_{ij}]
\end{equation}

where $\mathbf{W}_r$ is a learnable linear transformation matrix in the model, and $b_{ij}$ is obtained by calculating the union of the bounding boxes of target $i$ and target $j$. The bounding box of target $i$ is represented as $b_i = (x_{i1}, y_{i1}, x_{i2}, y_{i2})$, where $(x_{i1}, y_{i1})$ represents the horizontal and vertical coordinates of the upper left corner, and $(x_{i2}, y_{i2})$ represents the horizontal and vertical coordinates of the lower right corner. The bounding box of target $j$ is represented as $b_j = (x_{j1}, y_{j1}, x_{j2}, y_{j2})$. The union of bounding boxes $b_i$ and $b_j$, denoted as $b_{ij} = (x_{i1}, y_{j1}, x_{i2}, y_{j2})$, can be calculated by the following formulas:

\begin{equation}
	\begin{aligned}
		x_{i1} &= \min(x_{i1}, x_{j1}), \quad y_{i1} = \min(y_{i1}, y_{j1}) \\
		x_{i2} &= \max(x_{i2}, x_{j2}), \quad y_{i2} = \max(y_{i2}, y_{j2})
	\end{aligned}
\end{equation}

Here, $\max(\cdot)$ is used to determine the rightmost and lowest points of the union of two bounding boxes, and $\min(\cdot)$ is used to determine the leftmost and highest points of the union. Thus, this section obtains the initial graph of the dual graph $G_0 = \{V^0, E^0\}$.

Construction of the dual graph. To transform $G_0$ into the dual graph $G$, we convert the edges of $G_0$ into dual nodes and the nodes of $G_0$ into dual edges. Specifically, a node in the dual graph is formed by determining whether any two edges $x_{i \rightarrow j}$ and $x_{k \rightarrow l}$ connect the same target. If they do, there exists an edge $x$ connecting $x_{i \rightarrow j}$ and $x_{k \rightarrow l}$ in the dual graph, described by the following formulas:

\begin{equation}
	V = \{x_{i \rightarrow j} \mid x_{i \rightarrow j} \in E^0\}
\end{equation}

\begin{equation}
	E = \{x = (x_{i \rightarrow j}, x_{k \rightarrow l}) \mid x_{i \rightarrow j} \cap x_{k \rightarrow l} = x_i \in V^0\}
\end{equation}

where $x_{i \rightarrow j} \cap x_{k \rightarrow l} = x$ indicates that edges $x_{i \rightarrow j}$ and $x_{k \rightarrow l}$ connect the same target $x$. Thus, this section obtains the dual graph $G = \{V, E\}$.

\subsection{Question-Guided Intra-Type Message Passing}
\begin{figure*}[htbp]
	\centering
	\setlength{\belowcaptionskip}{-0.3cm}
	\includegraphics[width=170mm,height=58mm]{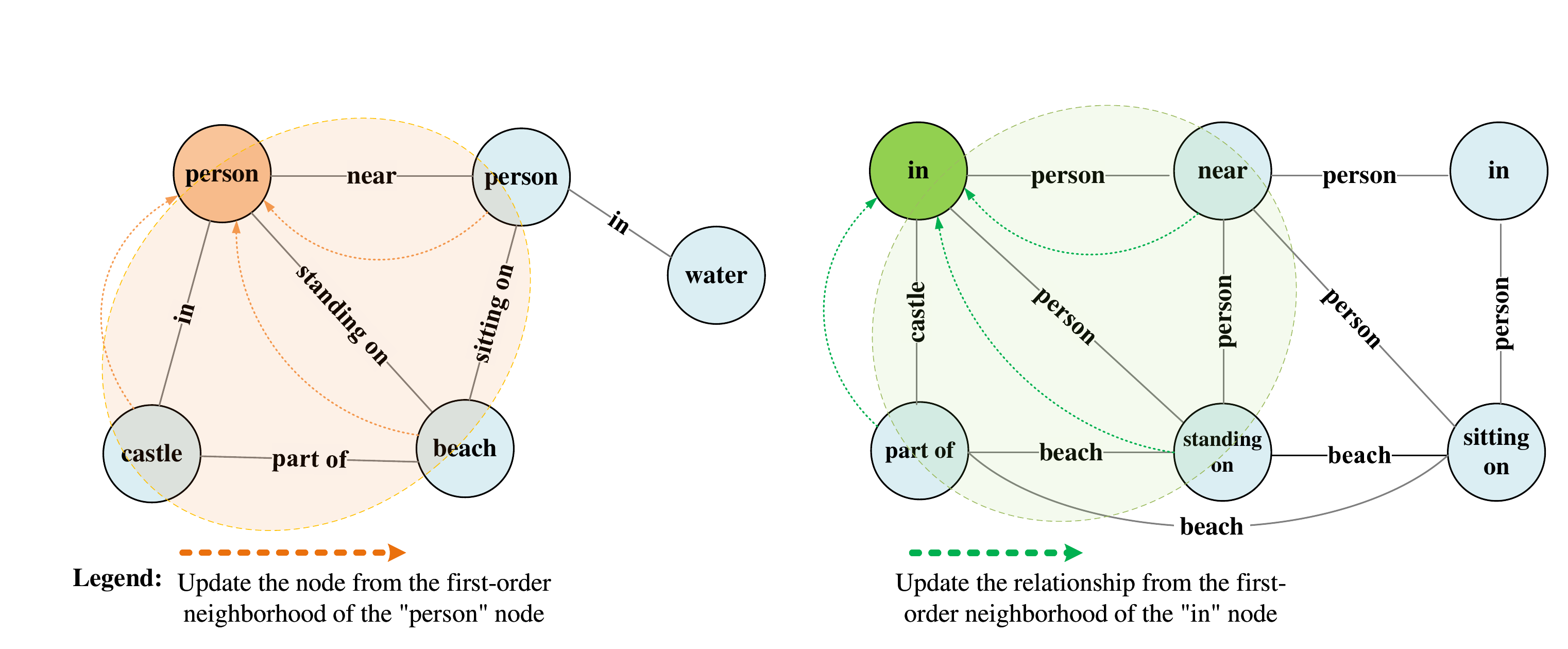}
	\caption{A schematic diagram of question-guided intra-type message passing reasoning, with the orange and green shades representing the first-order neighborhood of “person” and “in”.}
	\label{intra-type}
\end{figure*}

Intra-type message passing includes message updating within target classes and message updating within relationship classes. Specifically, during the message updating within target classes, the question guidance is used to focus on which target node within the first-order neighborhood of the target in the initial graph $G_0$ is more important for updating the target representation. During the message updating within relationship classes, the question guidance is used to focus on which relationship node within the first-order neighborhood of the relationship in the dual graph $G_d$ is more important for updating the relationship representation. As shown in Figure ~\ref{intra-type}, this section provides a schematic diagram of intra-type message passing reasoning.

Message updating within target classes. For the target $x_i$ in the initial graph $G_0$, this section passes the first-order neighborhood information related to the question to the target to update its feature representation. This process allows the model to learn the semantic context of the target object and improve its understanding of the question. Specifically, the first-order neighborhood of the target refers to all other targets $x_j \in N(x_i)$ directly connected to the target $x_i$. The target update formula is as follows:

\begin{equation}
	z_{x_i}^{(l+1)} = z_{x_i}^{(l)} + \sigma \left( \sum_{x_j \in N(x_i)} \alpha(x_j, q) \mathbf{W}_j z_{x_j}^{(l)} \right)
\end{equation}

\begin{equation}
	z_{x_i}^{(0)} = x_i
\end{equation}

where $z_{x_i}^{(l+1)}$ is the target representation of target $x_i$ at layer $(l+1)$, $\sigma(\cdot)$ denotes the non-linear activation function, $N(x_i)$ is the set of all targets in the first-order neighborhood of target $x_i$, $\mathbf{W}$ is the learnable weight matrix in the model, and $\alpha(x_j, q)$ is the attention weight indicating which target $x_j \in N(x_i)$ within the first-order neighborhood is more important for updating the target representation under the guidance of question $q$, with the specific calculation method as follows:

\begin{equation}
	\alpha(x_j, q) = \frac{\exp \left( \mathbf{W}^T \left[ z_{x_j}^{(l)}; c^{(l)} \right] \right)}{\sum_{x_j \in N(x_i)} \exp \left( \mathbf{W}^T \left[ z_{x_j}^{(l)}; c^{(l)} \right] \right)}
\end{equation}

where $z_{x_j}^{(l)}$ represents the target representation at layer $l$, $c^{(l)}$ represents the question instruction representation at layer $l$, and $\mathbf{W}^T$ represents the learnable weight matrix.

Message updating within relationship classes. For the target $x_{i \rightarrow j}$ in the dual graph $G_d$, the first-order neighborhood information related to the question is passed to this target to update its feature representation. This process allows the model to deeply learn the semantic context of the relationship object, thereby improving its understanding of the question. Specifically, the first-order neighborhood of the target refers to all other targets $x_{i \rightarrow k} \in N(x_{i \rightarrow j})$ directly connected to the target $x_{i \rightarrow j}$. The relationship update formula is as follows:

\begin{equation}
	e_{x_{i \rightarrow j}}^{(l+1)} = e_{x_{i \rightarrow j}}^{(l)} + \sigma \left( \sum_{x_{i \rightarrow k} \in N(x_{i \rightarrow j})} \alpha(x_{i \rightarrow k}, q) \mathbf{W}_g e_{x_{i \rightarrow k}}^{(l)} \right)
\end{equation}

\begin{equation}
	e_{x_{i \rightarrow j}}^{(0)} = x_{i \rightarrow j}
\end{equation}

where $e_{x_{i \rightarrow j}}^{(l+1)}$ is the target feature of $x_{i \rightarrow j}$ at layer $(l+1)$, $\mathbf{W}_g$ is the learnable weight matrix in the model, and $x_{i \rightarrow k} \in N(x_{i \rightarrow j})$ is the set of all targets within the first-order neighborhood of $x_{i \rightarrow j}$. $\alpha(x_{i \rightarrow k}, q)$ is the attention weight indicating which target $x_{i \rightarrow k} \in N(x_{i \rightarrow j})$ within the first-order neighborhood is more important for updating the relationship representation under the guidance of question $q$, with the specific calculation method as follows:

\begin{equation}
	\alpha(x_{i \rightarrow k}, q) = \frac{\exp \left( \mathbf{W}^T \left[ e_{x_{i \rightarrow k}}^{(l)}; c^{(l)} \right] \right)}{\sum_{x_{i \rightarrow k} \in N(x_{i \rightarrow k})} \exp \left( \mathbf{W}^T \left[ e_{x_{i \rightarrow k}}^{(l)}; c^{(l)} \right] \right)}
\end{equation}

where $e_{x_{i \rightarrow k}}^{(l)}$ represents the target representation at layer $l$, and $c^{(l)}$ represents the question instruction representation at layer $l$.

After obtaining the target features $z^{(L)}$ and relationship features $e^{(L)}$ from the dual graph, this section optimizes target classification and relationship classification using real target and relationship labels. Two linear classifiers are used to calculate the class probabilities $p_z$ and $p_e$ for targets and relationships, respectively:

\begin{equation}
	p_z = \text{softmax}(\mathbf{W}_z z^{(L)})
\end{equation}

\begin{equation}
	p_e = \text{softmax}(\mathbf{W}_e e^{(L)})
\end{equation}

where $\mathbf{W}_z$ and $\mathbf{W}_e$ represent the weight matrices of the target and relationship classifiers, respectively. This section combines two loss functions, including the binary cross-entropy loss $L_{\text{obj}}$ for optimizing target classification and the binary cross-entropy loss $L_{\text{rel}}$ for optimizing relationship classification:

\begin{equation}
	L_d = L_{\text{obj}} + L_{\text{rel}}
\end{equation}

Finally, this section inputs the updated target features $z^{(L)}$ within target classes and the updated relationship features $e^{(L)}$ within relationship classes into the fully connected function $FC(\cdot)$ to obtain the final feature vector $p_d$ for the final answer prediction. This vector contains visual representations related to the question within target and relationship classes, with the specific calculation formula as follows:

\begin{equation}
	p_d = \sigma \left( FC \left( \left[ z^{(L)}; e^{(L)} \right] \right) \right)
\end{equation}

\subsection{Heterogeneous Graph Construction Based on Static Relationships}
In this section, in order to obtain the representations between targets and relationships of different categories in videos, to achieve more accurate visual relationship context awareness, we construct a heterogeneous graph based on static relationships. This graph can effectively represent the different categories of targets and their interrelationships in the video. On the heterogeneous graph, a message passing reasoning model guided by questions is used to capture clues related to the question.

A heterogeneous graph is defined as $G_h = \{V, E, U, R\}$, where $V$ represents the set of target classes, $E$ represents the set of relationship classes, $U$ represents the set of target categories, and $R$ represents the set of relationship categories. For target nodes $u, v \in V$ and relationships $r_{u \rightarrow v} \in E$, node $v$ is assigned a target category $\phi(v) \in U$, and edge $r_{u \rightarrow v}$ is assigned a relationship category $\varphi(r_{u \rightarrow v}) \in R$. Heterogeneous graphs typically satisfy $|U| + |R| > 2$. In this section, $Y = (Y_o, Y_r)$ is defined to represent the set containing target categories $Y_o$ and relationship categories $Y_r$.

This section uses the Faster R-CNN model to extract target information from videos and construct the initial graph $G_0$. Specifically, for each detected target $u$, its visual features $v_u$, bounding box features $b_u$, and class label features $c_u$ are obtained. In the initial graph $G_0$, the node features of target $u$ are calculated by the following formula:

\begin{equation}
	x_u = \mathbf{W}_o [\mathbf{W}_v v_u; \mathbf{W}_b b_u; c_u]
\end{equation}

where $\mathbf{W}_o$, $\mathbf{W}_v$, and $\mathbf{W}_b$ are learnable linear transformation matrices in the model, and $[\cdot; \cdot]$ denotes the concatenation operation. The relationship between target $u$ and target $v$ is represented as $r_{u \rightarrow v}$, and its relationship features $x_{u \rightarrow v}$ in the initial graph $G_0$ can be defined as:

\begin{equation}
	x_{u \rightarrow v} = \mathbf{W}_r [x_u; x_v; b_{uv}]
\end{equation}

where $\mathbf{W}_r$ is a learnable linear transformation matrix in the model, and $b_{uv}$ is obtained by calculating the union of the bounding boxes of target $u$ and target $v$. Thus, this section obtains the initial graph of the heterogeneous graph $G_0 = \{V_0, E_0\}$.

Construction of the heterogeneous graph. To transform the initial graph $G_0$ into the heterogeneous graph $G_h$, this section uses the Faster R-CNN model to obtain the class logical values of all targets and assign classes to targets and relationships based on these class logical values.

Specifically, for the proposal of target $u$, it includes the class logical values obtained from the Faster R-CNN model, where each element of $p_u$ represents the class logical value of a target. This section uses a predefined function $\phi(\cdot)$ to calculate the target category $q_u \in \mathbb{R}^{|U|}$, which maps target categories to target categories, i.e., $\phi: Y_o \rightarrow U$. For example, for an image containing a scene of two people fencing, the detected targets $u$ and $v$ may belong to the "person" and "sword" categories, respectively. Through the $\phi(\cdot)$ function, this section can obtain the target categories $\phi(u) = \text{"Person (P)"}$, $\phi(v) = \text{"Non-Person (NP)"}$. The function $\phi(\cdot)$ can be a simple aggregation function such as $\text{sum}(\cdot)$ or $\text{mean}(\cdot)$.

The relationship category between two targets is predicted by a linear classifier on the relationship features $x_{u \rightarrow v}$ to generate the probability distribution of the relationship category $p_{u \rightarrow v} \in \mathbb{R}^{|R|}$, calculated as follows:

\begin{equation}
	p_{u \rightarrow v} = \text{softmax}(\mathbf{W}_{u \rightarrow v} x_{u \rightarrow v})
\end{equation}

where $\mathbf{W}_{u \rightarrow v}$ is the linear transformation matrix, and each element of $p_{u \rightarrow v}$ represents the probability of a relationship category. This section uses a predefined function $\varphi(\cdot)$ to infer the relationship category $q_{u \rightarrow v} \in \mathbb{R}^{|R|}$, which maps relationship categories to relationship categories, i.e., $\varphi: Y_r \rightarrow R$. In this section's work, three types of relationship categories are mainly considered: spatial, temporal, and contact. Therefore, this section constructs a heterogeneous graph with two types of target category nodes and three types of relationship category edges.

\subsection{Question-Guided Inter-Type Message Passing}
\begin{figure*}[htbp]
	\centering
	\setlength{\belowcaptionskip}{-0.3cm}
	\includegraphics[width=170mm,height=55mm]{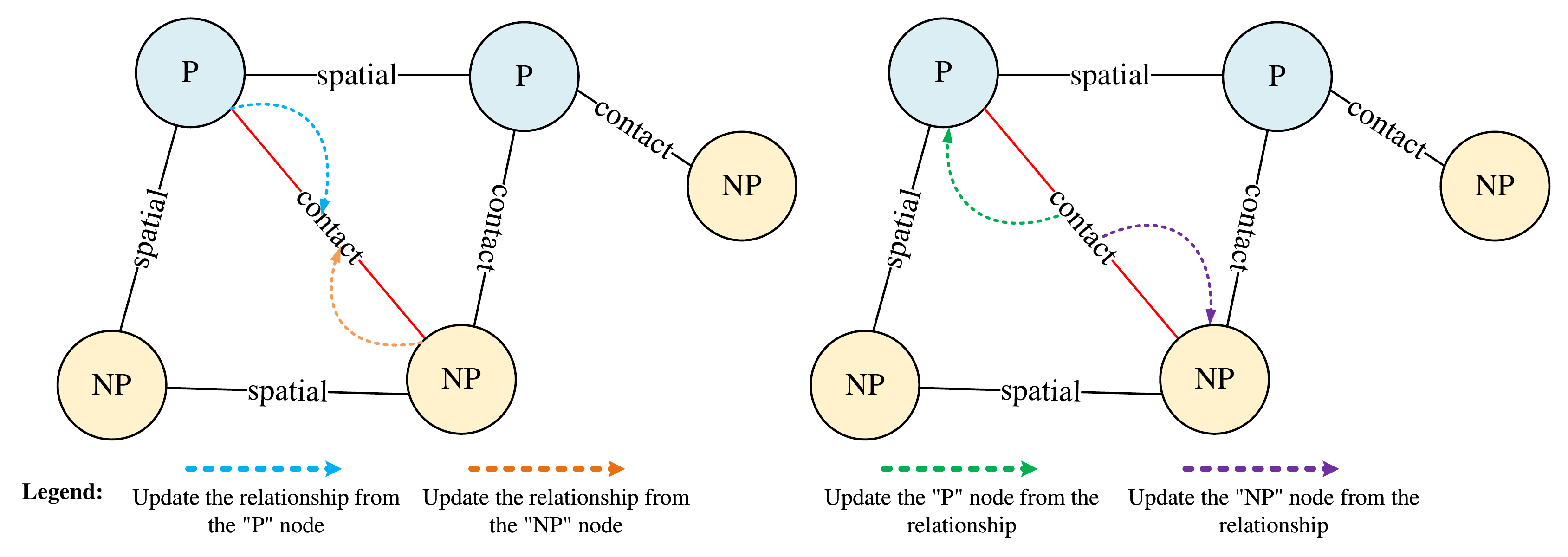}
	\caption{ A schematic diagram of inter-type message passing reasoning.}
	\label{inter-type}
\end{figure*}

Inter-type message passing includes message updating between target-relationship classes and relationship-target classes. Specifically, during the process of using messages between target-relationship classes to update relationships, question guidance is used to focus on which direction of node information in the heterogeneous graph $G_h$ is more important for updating the relationship representation. During the process of using messages between relationship-target classes to update targets, question guidance is used to focus on which relationship information within the first-order neighborhood of nodes in the heterogeneous graph $G_h$ is more important for updating the node representation. As shown in Figure ~\ref{inter-type}, this section provides a schematic diagram of the message passing reasoning between target-relationship classes.

Message updating between target-relationship classes. For target node $u$ and target node $v$ in heterogeneous graph $G_h$, there exists a relationship $x_{u \rightarrow v}$ with relationship category $\varphi(x_{u \rightarrow v})$, realizing the process of message passing between target classes to update relationships. Specifically, the above process includes two directions of message updating: from target node $u$ to relationship $x_{u \rightarrow v}$, and from target node $v$ to relationship $x_{u \rightarrow v}$. The process of message passing from node $u$ to node $v$ to update relationship $x_{u \rightarrow v}$ can be described as:

\begin{equation}
	e_{u \rightarrow v}^{(l+1)} = e_{u \rightarrow v}^{(l)} + \sigma \left( \alpha(u, q) \mathbf{W}_{\varphi(u \rightarrow v)}^{s2r} z_u^{(l)} + \beta(v, q) \mathbf{W}_{\varphi(u \rightarrow v)}^{s2r} z_v^{(l)} \right)
\end{equation}

\begin{equation}
	e_{u \rightarrow v}^{(0)} = x_{u \rightarrow v}, z_u^{(0)} = x_u
\end{equation}

where $e_{u \rightarrow v}^{(l+1)} \in \mathbb{R}^{d'}$ is the relationship representation of $x_{u \rightarrow v}$ at layer $(l+1)$, $\sigma(\cdot)$ is the non-linear activation function, $\mathbf{W}_{\varphi(u \rightarrow v)}^{s2r} \in \mathbb{R}^{d \times d'}$ represents the learnable weight matrix of relationship category $\varphi(x_{u \rightarrow v})$, and this matrix's calculation is inspired by the idea of basis decomposition. It can be obtained by calculating the first-order neighborhood information of node $u$ under different relationship categories, with specific calculations as follows:

\begin{equation}
	\mathbf{W}_{\varphi(u \rightarrow v)}^{s2r} = \sum_{i=1}^b \alpha_{\varphi(x_{u \rightarrow v})_i} \mathbf{B}_i
\end{equation}

where the superscript $s2r$ of $\mathbf{W}_{\varphi(u \rightarrow v)}$ indicates the direction of message passing from the subject to the relationship, $\mathbf{B} \in \mathbb{R}^{d \times d'}$ represents the trainable matrix of the $i$-th basis block, $b$ represents the number of basis blocks. $\alpha_{\varphi(x_{u \rightarrow v})_i}$ represents the trainable coefficient between relationship category $\varphi(x_{u \rightarrow v})$ and the basis block, capturing specific relationship category information. In Equation (4-20), $\alpha(u, q)$ is calculated by the similarity between $q$ and target node $u$, and $\beta(v, q)$ is calculated by the similarity between $q$ and target node $v$. The larger value represents that it is more important for updating the relationship, with specific calculation methods as follows:

\begin{equation}
	\alpha(u, q) = \frac{\exp \left( \mathbf{W}^T z_u^{(l)} \right)}{\exp \left( \mathbf{W}^T z_u^{(l)} \right) + \exp \left( \mathbf{W}^T c^{(l)} \right)}
\end{equation}

\begin{equation}
	\beta(v, q) = \frac{\exp \left( \mathbf{W}^T z_v^{(l)} \right)}{\exp \left( \mathbf{W}^T z_v^{(l)} \right) + \exp \left( \mathbf{W}^T c^{(l)} \right)}
\end{equation}

where $z_u^{(l)}$ and $z_v^{(l)}$ represent the target representations at layer $l$, $c^{(l)}$ represents the question instruction representation at layer $l$, $\mathbf{W}^T$ represents the learnable weight matrix, and $\exp(\cdot)$ is the normalization function.

Message updating between relationship-target classes. The update of target $u$ depends on the relationship representation $z_{u \rightarrow v}^{(l+1)}$ and $z_{v \rightarrow u}^{(l+1)}$ obtained during the message updating process between target-relationship classes, realizing the process of message passing between relationship classes to update targets. Specifically, given a node $u$ and its specified relationship category $t$'s first-order neighborhood information $N_t(u)$, to update its target node representation, the update calculation of target node $u$ at layer $(l+1)$ regarding relationship category $t$ is as follows:

\begin{equation}
	z_u^{(l+1)} = \sum_{v \in N_t(u)} \left\{ \alpha_{z_{u \rightarrow v}}(v, t) \mathbf{W}_{\varphi(x_{u \rightarrow v})}^{r2s} z_{u \rightarrow v}^{(l+1)} + \\
	 \alpha_{z_{v \rightarrow u}}(v, t) \mathbf{W}_{\varphi(x_{v \rightarrow u})}^{r2s} z_{v \rightarrow u}^{(l+1)} \right\}
\end{equation}

where the superscript $r2s$ of $\mathbf{W}_{\varphi(x_{u \rightarrow v})}$ indicates the direction of message passing from the relationship to the subject, $\mathbf{W}_{\varphi(x_{u \rightarrow v})} \in \mathbb{R}^{d \times d}$ and $\mathbf{W}_{\varphi(x_{v \rightarrow u})} \in \mathbb{R}^{d \times d}$ are the weight matrices of relationship category $\varphi(x_{u \rightarrow v})$, representing the messages passed from relationship $x_{u \rightarrow v}$ to target $x_u$ and from relationship $x_{v \rightarrow u}$ to target $x_u$, respectively.
The values of $\alpha_{r2s}(v,t)$ and $\alpha_{2o}(v,t)$ represent the importance of target node $u$ with respect to target node $v$ in the first-order neighborhood under the condition of relationship category $t$. $\alpha_{2o}(v,t)$ is calculated by the similarity between $q$ and target node $v$, indicating the importance of updating the relationship:

\begin{equation}
	\alpha_{2o}(v,t) = \frac{\exp \left( \mathbf{W}^T z_v^{(l)} \right)}{\sum_{t \in N_t(u)} \exp \left( \mathbf{W}^T z_{t \rightarrow u}^{(l)} \right)}
\end{equation}

Here, $\alpha_{r2s}(v,t)$ and $\alpha_{2o}(v,t)$ values respectively indicate the importance of the relationship from node $u$ to node $v$ and from node $v$ to node $u$ under the condition of relationship category $t$, which information is more important for updating the node. This section aggregates all target features based on specific relationship categories to obtain the final target feature $z_u^{(l+1)}$:

\begin{equation}
	z_u^{(l+1)} = z_u^{(l)} + \frac{1}{|R|} \sum_{i=1}^{|R|} \sigma \left( \alpha_{u \rightarrow i}^{(l+1)} \right)
\end{equation}

After obtaining the target features $z^{(L)}$ and relationship features $e^{(L)}$ from the heterogeneous graph, this section optimizes target classification and relationship classification using real target and relationship labels. Two linear classifiers are used to calculate the class probabilities $p_z$ and $p_e$ for targets and relationships, respectively:

\begin{equation}
	p_z = \text{softmax}(\mathbf{W}_z z^{(L)})
\end{equation}

\begin{equation}
	p_e = \text{softmax}(\mathbf{W}_e e^{(L)})
\end{equation}

where $\mathbf{W}_z$ and $\mathbf{W}_e$ represent the weight matrices of the target and relationship classifiers, respectively. This section combines two loss functions: the binary cross-entropy loss $L_{\text{obj}}$ for optimizing target classification and the binary cross-entropy loss $L_{\text{rel}}$ for optimizing relationship classification:

\begin{equation}
	L_h = L_{\text{obj}} + L_{\text{rel}}
\end{equation}

Finally, this section inputs the updated node features $z^{(L)}$ from relationship-target classes and the updated relationship features $e^{(L)}$ from target-relationship classes into the $FC(\cdot)$ function to obtain the final feature vector $p_h$ for the final answer prediction. This vector contains visual representations related to the question between target-relationship classes and relationship-target classes, with the specific calculation formula as follows:

\begin{equation}
	p_h = \sigma \left( FC \left( \left[ z^{(L)}; e^{(L)} \right] \right) \right)
\end{equation}

\subsection{Answer Generation}
This section employs a common classification method to construct a single-hop attention model as a classifier. For open-ended questions, answers are generated from the vocabulary of the dataset. Specifically, in Equation (4-1), this section uses a Bi-LSTM to encode the question $Q$ into a vector $q$. Then, single-hop attention $\lambda_d$ is used to aggregate visual information from the dual graph related to the question, and single-hop attention $\gamma_h$ is used to aggregate visual information from the heterogeneous graph related to the question to predict the answer:

\begin{equation}
	\begin{aligned}
		\lambda_d &= \text{softmax}(p_d \odot \mathbf{W}_q q), \quad \gamma_h = \text{softmax}(p_h \odot \mathbf{W}_q q) \\
		y &= \mathbf{W}_o \sigma \left( (\lambda_d p_d + \gamma_h p_h)^T q \right)
	\end{aligned}
\end{equation}

where $\mathbf{W}_q$ and $\mathbf{W}_o$ are trainable parameters. During the training process, the output scores $y$ of the answer classifier are subject to cross-entropy classification loss $L_s$. For multiple-choice questions, the correct answer $\hat{a}$ is selected from the candidate answers using class-level cues $p_d$ and inter-type cues $p_h$ related to the question. Specifically, the similarity between the candidate answer $a$ and the class-level cues $p_d$ and inter-type cues $p_h$ is calculated to generate a vector $z = \{z_i\}_{i=1}^5$ containing five confidence scores. The candidate answer with the highest confidence score is selected as the predicted answer $\hat{a}$, with specific calculations as follows:

\begin{equation}
	\begin{aligned}
		z &= \text{softmax} \left( (\lambda_d p_d + \gamma_h p_h)^T a \right) \\
		\hat{a} &= \arg \max_{i \in [1,5]} (z_i)
	\end{aligned}
\end{equation}

To train the entire video question answering model, we combine the optimization of intra-type message passing loss $L_d$, inter-type message passing loss $L_h$, and question answering loss $L_a$ as the overall loss $L = L_d + L_h + L_a$. By minimizing the loss function, the model's performance in target and relationship classification and question answering can be simultaneously improved, ensuring more accurate question answering.

\section{Experiments}
\label{sec:Experiments}
To demonstrate the capabilities of TAMP, our experimental approach is as follows: First, we introduce the dataset and the evaluation metrics used. Then, we methodically present the experimental results to prove TAMP's reasoning capabilities. Finally, the visualization of the reasoning process is shown.
\subsection{Dataset and Metrics}
\textbf{Dataset.}  The ANetQA\cite{yu2023anetqa} dataset is a large-scale video question answering benchmark dataset created by the School of Computer Science and Technology at Hangzhou Dianzi University. It is designed to support fine-grained compositional reasoning on untrimmed videos. The construction of the ANetQA dataset is based on the ActivityNet\cite{caba2015activitynet} dataset, and it generates 1.4 billion imbalanced and 13.4K balanced question-answer pairs through an automated process. These pairs are automatically derived from annotated video scene graphs, reflecting the fine-grained semantics, spatiotemporal scene graphs, and diverse question-answer templates of the videos.

The Next-QA\cite{xiao2021next} dataset was developed by researchers from the Department of Computer Science at the National University of Singapore. It aims to elevate the challenge of video question answering from descriptive questions to explanatory questions, particularly in the areas of causal action reasoning and temporal action reasoning. The dataset consists of 5.4K carefully constructed videos, containing approximately 47.7K multiple-choice question-answer pairs and around 52K open-ended question-answer pairs. These videos capture rich object interactions and multi-step reasoning from everyday activities. 

\textbf{Metrics.} The TAMP's answering capabilities are evaluated using several key metrics: Accuracy,  Recall@K (R@K) and mean Recall@K (mR@K).

\textbf{Implementation details.}
For visual features, this chapter employs a Feature Pyramid Network\cite{lee2023afi} and Faster R-CNN as the object detector to extract objects from video frames. By default, the non-maximum suppression value for each object category is set to 0.5, and the top 80 objects are selected to construct the initial graph. For textual features, this chapter initializes the question word embeddings from the pre-trained GloVe\cite{pennington2014glove} model. During training, the model maintains an exponential moving average of the parameters, which smooths the parameter updates, reduces fluctuations in the training process, and helps the model converge more quickly to a more stable state.

During training, the network layers before the ROIAlign layer are frozen, and the remaining network layers are optimized using object and relationship classification loss functions. This chapter adopts the Adam\cite{kingma2014adam} optimizer to minimize the objective function, with an initial learning rate set to 0.008, a batch size of 6, and a weight decay of 1.0e-05. For intra-type and inter-type message passing reasoning, the method performs 3 iterations, with the iteration parameter set to 3. All experiments are conducted on the PyTorch platform and accelerated using CUDA on an NVIDIA TESLA V100 GPU.

\subsection{Performance on ANetQA}

On the ANetQA dataset, this chapter selected three state-of-the-art models for comparison, namely HCRN\cite{le2020hierarchical}, ClipBERT\cite{lei2021less}, and All-in-One\cite{wang2023all}. HCRN introduces a reusable conditional relational network to integrate multi-level motion, question, and visual features for reasoning. ClipBERT and All-in-One are two Transformer-based video question answering models that combine visual language pre-training on large-scale corpora. ClipBERT was retrained on a vast number of image-text pairs, achieving end-to-end learning through a sparse sampling mechanism. All-in-One first uses raw videos and texts as inputs to an end-to-end video-language pre-training model, directly pre-trained on a large-scale video-text corpus.

Table~\ref{table1} shows the comparative experimental results of our proposed model TAMP with the aforementioned advanced models HCRN, ClipBERT, and All-in-One. In addition to overall accuracy, we also present the accuracy of different types of questions under various classification methods, namely question structure, question semantics, reasoning skills, and answer types. Among the question structure categories, query-type questions are the most challenging because they contain a rich set of answers. However, TAMP achieved the best results in the query category because it excels at identifying fine-grained visual representations of targets and relationships. The remaining compare, choose, verify, and logic types of questions are relatively simple because their answers are selected from a finite set of options. Choose questions require more reasoning steps, hence the reasoning results of HCRN, ClipBERT, and other models are slightly lower than those of the TAMP model. TAMP utilizes question-guided intra-type and inter-type message passing for multi-step iterative reasoning, achieving an accuracy of 56.10

\begin{table}[h]
	\centering
	\caption{The experiment results of ANetQA dataset}
	\begin{tabular}{ccccc}
		\hline
		Question Category & HCRN & ClipBERT& All-in-One & Ours \\
		\hline
		query & 21.30 & 23.93 & 25.10 & \textbf{26.75} \\
		compare & 55.66 & 55.62 & 54.41 & \textbf{56.10} \\
		choose & 63.97 & 69.51 & 70.39 & 68.32 \\
		verify & 68.56 & 72.57 & 72.35 & 72.31 \\
		logic & 78.70 & 80.06 & 80.58 & \textbf{81.53} \\
		\hline object & 55.99 & 58.69 & 59.81 & \textbf{60.14} \\
		relationship & 39.65 & 40.19 & 40.78 & \textbf{40.85} \\
		attribute & 35.80 & 39.71 & 40.14 & 36.21 \\
		action & 72.50 & 74.96 & 74.39 & 73.39 \\
		\hline object-relationship & 35.17 & 37.66 & 38.42 & \textbf{41.43} \\
		object-attribute & 40.95 & 43.72 & 44.33 & 43.49 \\
		duration-comparison & 49.90 & 49.98 & 51.65 & 50.77 \\
		exist & 71.20 & 74.51 & 74.49 & \textbf{77.34} \\
		sequencing & 31.70 & 34.19 & 35.27 & 35.87 \\
		superlative & 47.46 & 49.55 & 50.14 & 49.67 \\
		\hline binary & 64.36 & 66.19 & 65.65 & \textbf{67.53} \\
		open & 29.95 & 33.17 & 34.33 & \textbf{35.31} \\
		\hline Overall & 41.15 & 43.92 & 44.53 & \textbf{45.92} \\
		\hline
	\end{tabular}\label{table1}
\end{table}
In the classification of question semantics, TAMP achieved the best results in the object and relationship categories, precisely because TAMP excels at representing and recognizing targets and relationships. Questions oriented towards attribute types are the most challenging, as they require the model to have finer-grained attribute representation and recognition capabilities for video content. In this regard, the All-in-One model leveraged knowledge from a large-scale video-text corpus to achieve an accuracy rate of $40.14\%$, nearly $4\%$ higher than our model. As for action-type questions, ClipBERT effectively represented video motion features through a sparse sampling mechanism, achieving an accuracy rate of $74.96\%$, $1.5\%$ higher than our model. Based on the analysis of the above results, future work of this method will verify whether the effective representation of attributes and motion has a positive correlation with the performance of TAMP.

In the classification of reasoning skills, each type of question requires the model to have multi-step reasoning capabilities to answer the questions. TAMP utilized intra-type message passing models to capture fine-grained target and relationship representations centered on targets and relationships, thereby achieving an accuracy rate of $41.43\%$ on object-relationship type questions. At the same time, by using question-guided multi-step intra-type and inter-type message passing reasoning, fine-grained target and relationship representations related to the question were obtained, enabling TAMP to achieve good results on exist and sequencing type questions. The actions in the ANetQA dataset are described in natural language, which is more difficult for models without motion recognition to understand. In the classification of answer types, TAMP achieved accuracy rates of $67.53\%$ and $35.31\%$, respectively, fully demonstrating the model's effective modeling and reasoning of relationships in videos, and making full use of static relationships in videos to enhance the model's reasoning capabilities.

\subsection{Performance on Next-QA}
To verify the generalization capability of the method proposed in this chapter, comparative experiments were conducted on the Next-QA dataset. Next, a brief introduction to the models compared in the experiment is provided.

HCRN\cite{le2020hierarchical}: Proposed a conditional relational network (CRN) that can transform the input object set into an encoded output object set representing input relationships. By stacking CRNs, a hierarchical network is constructed to effectively handle long video question answering.

HME\cite{fan2019heterogeneous}: Enhanced multimodal attention through a heterogeneous memory network, integrating appearance and motion features, and designed a multimodal fusion layer. Multi-step reasoning is performed through a self-updating attention mechanism to infer answers.

HGA\cite{jiang2020reasoning}: Proposed a graph construction to represent videos and questions, combining modular collaborative attention models and graph convolutional networks for cross-modal fusion and alignment, achieving cross-modal reasoning.

HQGA\cite{xiao2022video}: Modeled videos as conditional graph hierarchies, integrating visual facts of different granularities in a hierarchical manner, and combining questions at each level to locate relevant video elements.

ATP\cite{buch2022revisiting}: Proposed a multimodal model based on image-level understanding for video language analysis, effectively integrating ATP into a complete video-level temporal model to improve the efficiency and accuracy of video question answering models.

VGT\cite{xiao2022video}: Proposed a dynamic graph transformer module to explicitly capture visual objects, their relationships, and dynamics for complex spatiotemporal reasoning, improving visual relationship reasoning in video question answering tasks.

MIST\cite{gao2023mist}: Used a cascade of segment and region selection modules to adaptively select frames and image regions closely related to the question, supporting multi-event reasoning through iterative selection and attention mechanisms.

ATM\cite{chen2023atm}: Utilized optical flow to capture long-term temporal reasoning and perform action-centric contrastive learning, focusing on extracting action phrases from questions to learn rich cross-modal embeddings.

GF\cite{bai2023glance}: Generated dynamic event memories through an encoder-decoder, designing an unsupervised memory generation method. The generated memories were reorganized into memory prompts, serving as a bridge between questions and video content.

IGV\cite{li2022invariant}: Proposed a new learning framework, invariance localization for video question answering, to locate scenes with invariant causal relationships with answers, shielding the negative impact of spurious correlations, and enhancing the model's reasoning capabilities.

In the latest research progress in the field of video question answering, the TAMP method proposed in this chapter has shown significant advantages in performance. As shown in Tables~\ref{table2} and~\ref{table3}, compared with the current state-of-the-art methods, experimental results demonstrate that TAMP's performance on the Next-QA dataset is particularly outstanding. This method adopts a consistent object feature extraction strategy and does not rely on any additional external data pre-training. On the test and validation sets of the Next-QA dataset, TAMP has surpassed existing excellent methods, achieving overall accuracy rates of $60.42\%$ and $50.85\%$, respectively.

\begin{table}[h]
	\centering
	\caption{Results of multi-choice QA on test set of Next-QA dataset}
	\begin{tabular}{ccccc}
		\hline
		\textbf{Model} & \textbf{ACC\%@C} & \textbf{ACC\%@T} & \textbf{ACC\%@D} & \textbf{ACC\%@ALL} \\
		\hline
		HCRN & 47.07 & 49.27 & 54.02 & 48.89 \\
		HME & 46.76 & 48.89 & 57.37 & 49.16 \\
		HGA & 48.13 & 49.08 & 57.79 & 50.01 \\
		HQGA & 49.04 & 52.28 & 59.43 & 51.75 \\
		IGV & 48.56 & 51.67 & 59.64 & 51.34 \\
		VGT & 51.62 & 51.94 & 63.65 & 53.68 \\
		ATM & 55.31 & 55.55 & 65.34 & 57.03 \\
		Ours & \textbf{58.14} & \textbf{58.02} & \textbf{68.86} & \textbf{60.42} \\
		\hline
	\end{tabular}\label{table2}
\end{table}

The effectiveness of TAMP's experimental results is demonstrated across different question types, including the accuracy rate on causal questions (ACC\%@C), temporal questions (ACC\%@T), and descriptive questions (ACC\%@D). As shown in Table ~\ref{table2}, TAMP performed excellently on causal reasoning questions, achieving an accuracy rate of $58.14\%$. This is because it utilizes question guidance to conduct message passing between targets of the same and different categories, capturing fine-grained target and relationship information related to the question. After multiple iterations of message passing, it effectively inferred the causal information in the video for answering the question. In Table~\ref{table3}, TAMP achieved the second-best results on descriptive questions (indicated by underlined results in the table), and the best results on causal questions, temporal questions, and overall accuracy, fully demonstrating the stability and generalization of the TAMP method.

\begin{table}[h]
	\centering
	\caption{Results of multi-choice QA on validation set of Next-QA dataset}
	\begin{tabular}{ccccc}
		\hline
		\textbf{Model} & \textbf{ACC\%@C} & \textbf{ACC\%@T} & \textbf{ACC\%@D} & \textbf{ACC\%@ALL} \\
		\hline
		HCRN & 45.91 & 49.26 & 53.67 & 48.20 \\
		HME & 46.18 & 48.20 & 58.30 & 48.72 \\
		HGA & 46.26 & 50.74 & 59.33 & 49.74 \\
		HQGA & 48.48 & 51.24 & 61.65 & 51.42 \\
		ATP & 53.10 & 50.20 & 66.80 & 54.30 \\
		VGT & 52.28 & 55.09 & 64.09 & 55.02 \\
		MIST & 54.62 & 56.64 & 66.92 & 57.18 \\
		ATM & 56.04 & 58.44 & 65.38 & 58.27 \\
		GF & 56.93 & 57.07 & 70.53 & 58.83 \\
		Ours & \textbf{59.17} & \textbf{60.34} & \textbf{69.76} & \textbf{60.85} \\
		\hline
	\end{tabular}\label{table3}
\end{table}

The success of TAMP is attributed to its three core components: the question instruction module, intra-type message passing, and inter-type message passing modules. The collaborative work of these components enables TAMP to model the representations of targets and relationships finely and perform effective reasoning. Compared to traditional methods based on coarse-grained video clipping and alignment with question answering, TAMP can capture higher-level semantic feature representations, demonstrating its precision and robustness in handling relationship-intensive tasks. TAMP's advantages in these tasks surpass existing methods, highlighting its excellent capability in relationship modeling.

\subsection{Performance On Ablation Study}
We conducted ablation studies to verify the effectiveness of intra-type and inter-type message passing models, respectively. This method uses Recall@K (R@K) and mean Recall@K (mR@K) to evaluate the target and relationship recognition results of intra-type and inter-type message passing reasoning models, while also recording the accuracy of answer questions. The experimental results of the ablation study are shown in Table~\ref{table4}. The experiment includes three ablation models, namely intra-type message passing reasoning model ($\mathrm{TAMP}_{intra}$), inter-type message passing reasoning model ($\mathrm{TAMP}_{inter}$), and intra-type and inter-type message passing reasoning model ($\mathrm{TAMP}_{intra+inter}$). The $\mathrm{TAMP}_{intra}$ model focuses on the question guidance to focus on the target class and relationship class for question answering, while the $\mathrm{TAMP}_{inter}$ model combines intra-type and inter-type message passing for question answering. From the experimental results, the target recognition of $\mathrm{TAMP}_{intra+inter}$ achieved $33.73\%$ and $38.32\%$ in R@50 and R@100, respectively, and the relationship recognition achieved $14.79\%$ and $16.68\%$ in mR@50 and mR@100, respectively. The answer question accuracy is $45.92\%$, which is better than the target recognition and question results of both $\mathrm{TAMP}_{intra}$ and $\mathrm{TAMP}_{inter}$ models. At the same time, we found that $\mathrm{TAMP}_{intra}$ and $\mathrm{TAMP}_{inter}$ each have their unique advantages, and both are very important for answering questions. However, they also concluded that the more accurate the target and relationship recognition results in the video, the higher the accuracy of the answer. In summary, each model has played a key role in enhancing the performance of TAMP, and their collaborative effect has significantly improved the performance of TAMP.

\begin{figure}[htbp]
	\centering
	\setlength{\belowcaptionskip}{-0.2cm}
	\includegraphics[width=90mm,height=35mm]{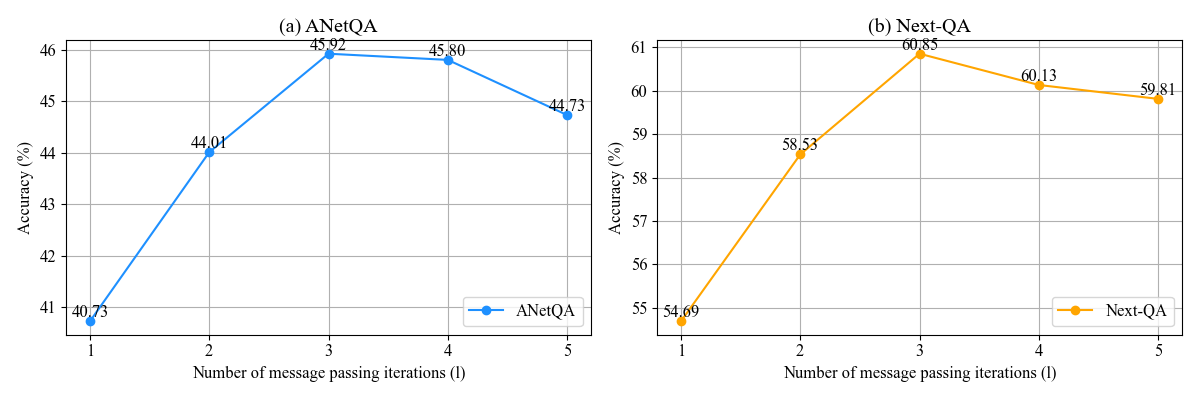}
	\caption{Ablation results of the number of iterations $l$ in intra- and inter-type message passing reasoning model}
	\label{train_loss}
\end{figure}

\begin{table}[h]
	\centering
	\caption{The ablation results of intra- and inter-type message passing reasoning models on ANetQA dataset}
	\scalebox{0.86}{
	\begin{tabular}{cccccc}
		\hline
		\textbf{Model} & \textbf{R@50} & \textbf{R@100} & \textbf{mR@50} & \textbf{mR@100} & \textbf{Accuracy\%} \\
		\hline
		$\mathrm{TAMP}_{intra}$ & 32.53 & 37.68 & 12.26 & 14.24 & 44.32 \\
		$\mathrm{TAMP}_{inter}$ & 32.19 & 36.54 & 13.71 & 15.83 & 44.06 \\
		$\mathrm{TAMP}_{intra+inter}$ & \textbf{33.73} & \textbf{38.32} & \textbf{14.79} & \textbf{16.68} & \textbf{45.92} \\
		\hline
	\end{tabular}}\label{table4}
\end{table}
To select the optimal number of message passing iterations for the model, this method conducted ablation experiments on the ANetQA and Next-QA datasets regarding the number of message passing iterations $l$ in the intra-type and inter-type message passing reasoning model of the TAMP model. As shown in Figure~\ref{train_loss}, the x-axis represents the number of message passing iterations, and the y-axis indicates the model's accuracy in answering questions ($\%$). On the ANetQA dataset, the accuracy steadily increased from $40.73\%$ in the first iteration to $45.92\%$ in the third iteration, and then the accuracy was $45.80\%$ and $44.73\%$ in the fourth and fifth iterations, respectively. For the Next-QA dataset, the accuracy started at $54.69\%$, and with the increase in the number of iterations, it reached the highest point of $60.85\%$ in the third iteration, then slightly decreased, and finally the accuracy was $59.81\%$ in the fifth iteration.

The experimental results indicate that when the number of message passing iterations is 3, the TAMP model achieved the highest accuracy on both datasets, suggesting that three iterations of message passing is a suitable choice that balances performance and computational efficiency. Additionally, from the training results on the ANetQA and Next-QA datasets, the model demonstrated higher stability after reaching a certain amount of training.

\subsection{Qualitative Analysis}

\begin{figure}[htbp]
	\centering
	\setlength{\belowcaptionskip}{-0.2cm}
	\includegraphics[width=90mm,height=60mm]{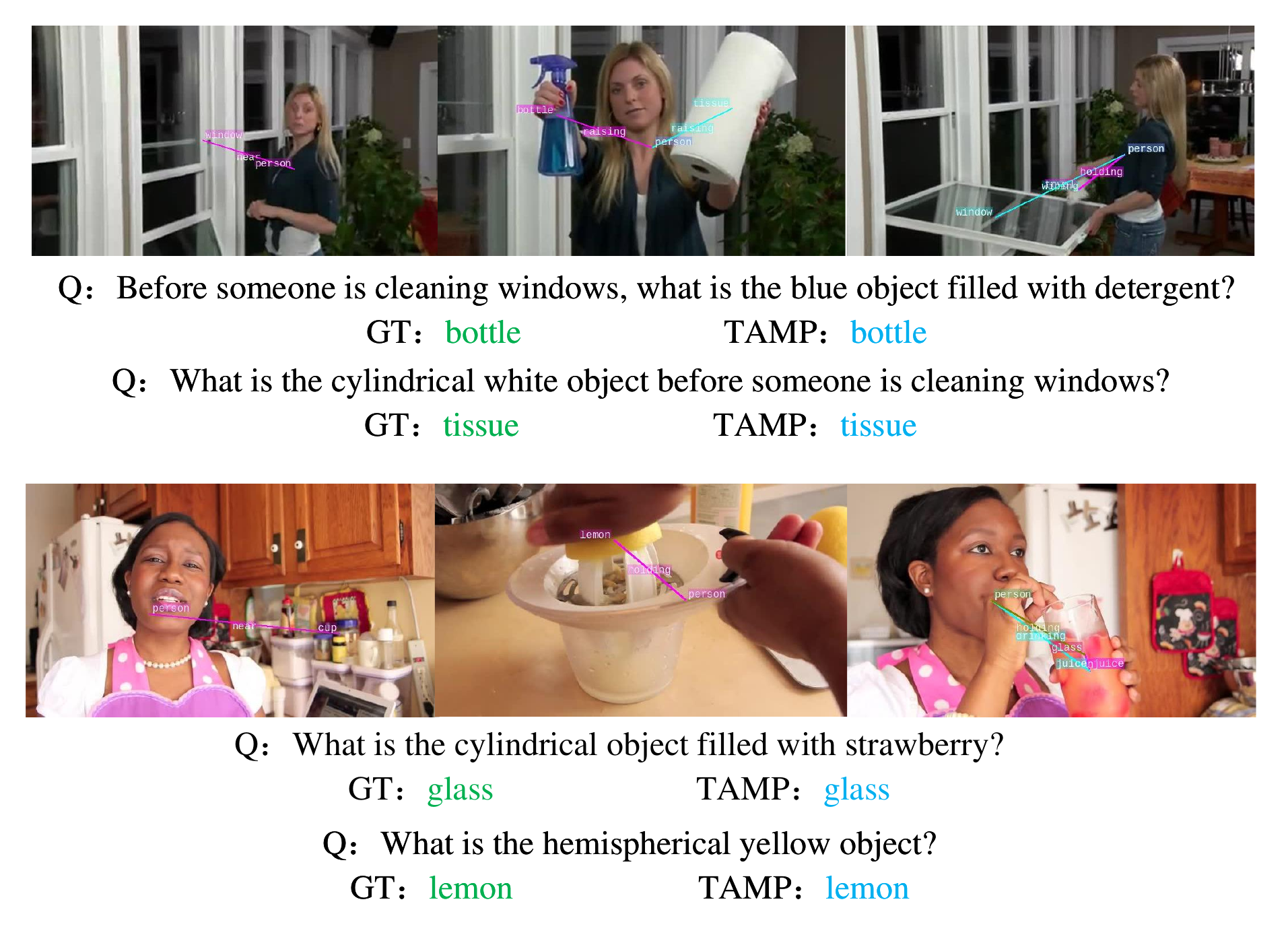}
	\caption{Visualization of predicted correct results on the ANetQA dataset.}
	\label{example1}
\end{figure}

\begin{figure}[htbp]
	\centering
	\setlength{\belowcaptionskip}{-0.2cm}
	\includegraphics[width=90mm,height=30mm]{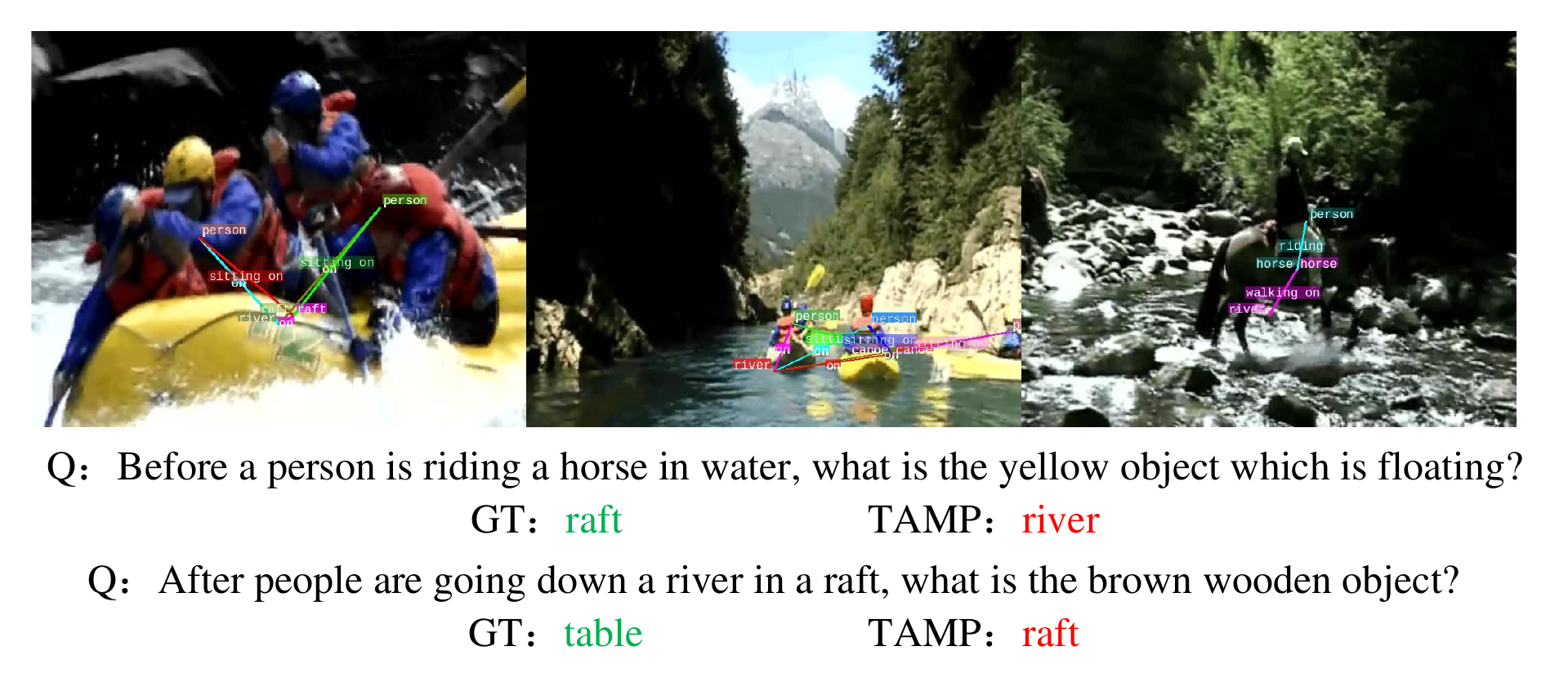}
	\caption{Visualization of predicted incorrect results on the Next-QA dataset.}
	\label{example2}
\end{figure}

This study aims to evaluate the effectiveness of the proposed intra-type and inter-type message passing reasoning method based on static relationships in the video question answering task through visualization results. As shown in Figures~\ref{example1} and ~\ref{example2}, visualization experiments were conducted on the ANetQA [71] dataset, where each video is represented by three frames, and the static relationships related to the question answers are highlighted in the figures. The ground-truth answers are displayed in green, correct model predictions are shown in blue, and incorrect model predictions are marked in red.

In Figure~\ref{example1}, the model accurately identifies the blue object as a “bottle” and the white cylindrical object as “tissue,” consistent with the ground-truth answers. This demonstrates the model's potential in handling video question answering tasks that involve fine-grained static relationships. Similarly, when dealing with a cylindrical object containing strawberries, the model correctly identifies it as a “glass,” which matches the ground-truth answer. Additionally, when recognizing a hemispherical yellow object, the model predicts “lemon,” which is also the correct answer. These results indicate that the model effectively captures the contextual clues within the object class and relationship class through intra-type message passing. Meanwhile, inter-type message passing obtains contextual clues between objects and relationships, as well as between relationships and objects. The integration of intra-type and inter-type clues in this chapter significantly enhances the accuracy of question reasoning.

In Figure~\ref{example2}, the model failed to accurately identify the yellow floating object as a "raft," instead incorrectly predicting it as a "river." Additionally, when recognizing a brown wooden object, the model predicted "raft," while the ground-truth answer was "table." These errors highlight the challenges the model faces in handling dynamic scenes and distinguishing between similar objects.

Future work will focus on integrating object attributes and events more deeply into the existing framework to enhance the model's understanding and reasoning capabilities regarding video content. By incorporating richer semantic information, we expect the model to achieve higher-level semantic feature representations and more accurate reasoning when dealing with complex video question answering tasks.

\section{Conclusion}
This paper delves into the research question of how to effectively model and reason about relationships within videos, and how to leverage static relationships within videos to enhance the model's reasoning capabilities. We propose a method based on intra-type and inter-type message passing reasoning that is grounded in static relationships. By generating question instructions, constructing dual graphs, and facilitating intra-type message passing, our approach effectively captures clues related to the question within both target and relationship classes. Furthermore, this chapter explores an inter-type message passing reasoning model that constructs heterogeneous graphs and passes messages between classes, effectively capturing clues related to the question between target-relationship classes and relationship-target classes, thereby enhancing the model's deep understanding of static relationships in videos.

In the experimental section, we conducted thorough experimental analyses on the ANetQA and Next-QA datasets, validating the effectiveness of the methods proposed in this chapter. The experimental results demonstrate the importance of both intra-type and inter-type message passing. Specifically, the intra-type clues constructed based on dual graphs and the inter-type clues constructed based on heterogeneous graphs each have their unique advantages, and both are crucial for answering questions. Additionally, we observed a clear conclusion: the higher the accuracy of target and relationship recognition in videos, the higher the accuracy of question answering.

In future research, we can integrate target attributes and events into the existing framework to deepen the model's cognition and reasoning capabilities regarding video content. This will help further improve the model's performance in complex tasks such as video question answering, providing new perspectives and methods for the field of video understanding and reasoning.


\bibliographystyle{IEEEtran}
\bibliography{tamp.bib}   
\makeatletter
\begin{IEEEbiography}[{\includegraphics[width=1in,height=1.25in,clip,keepaspectratio]{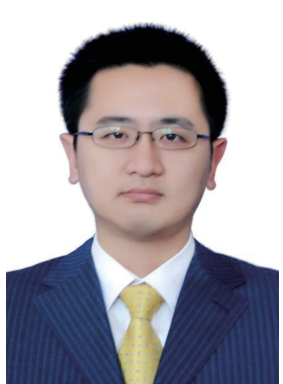}}]{Guanglu Sun}
	(Member, IEEE) received his Ph.D. in Computer Application Technology from Harbin Institute of Technology in 2008. From 2014 to 2015, he was a visiting scholar at Northwestern University, USA. He is currently a professor and the Director of the Center of Information Security and Intelligent Technology at Harbin University of Science and Technology. His research interests include computer networks and security, machine learning, and intelligent information processing.
\end{IEEEbiography}

\begin{IEEEbiography}[{\includegraphics[width=1in,height=1.25in,clip,keepaspectratio]{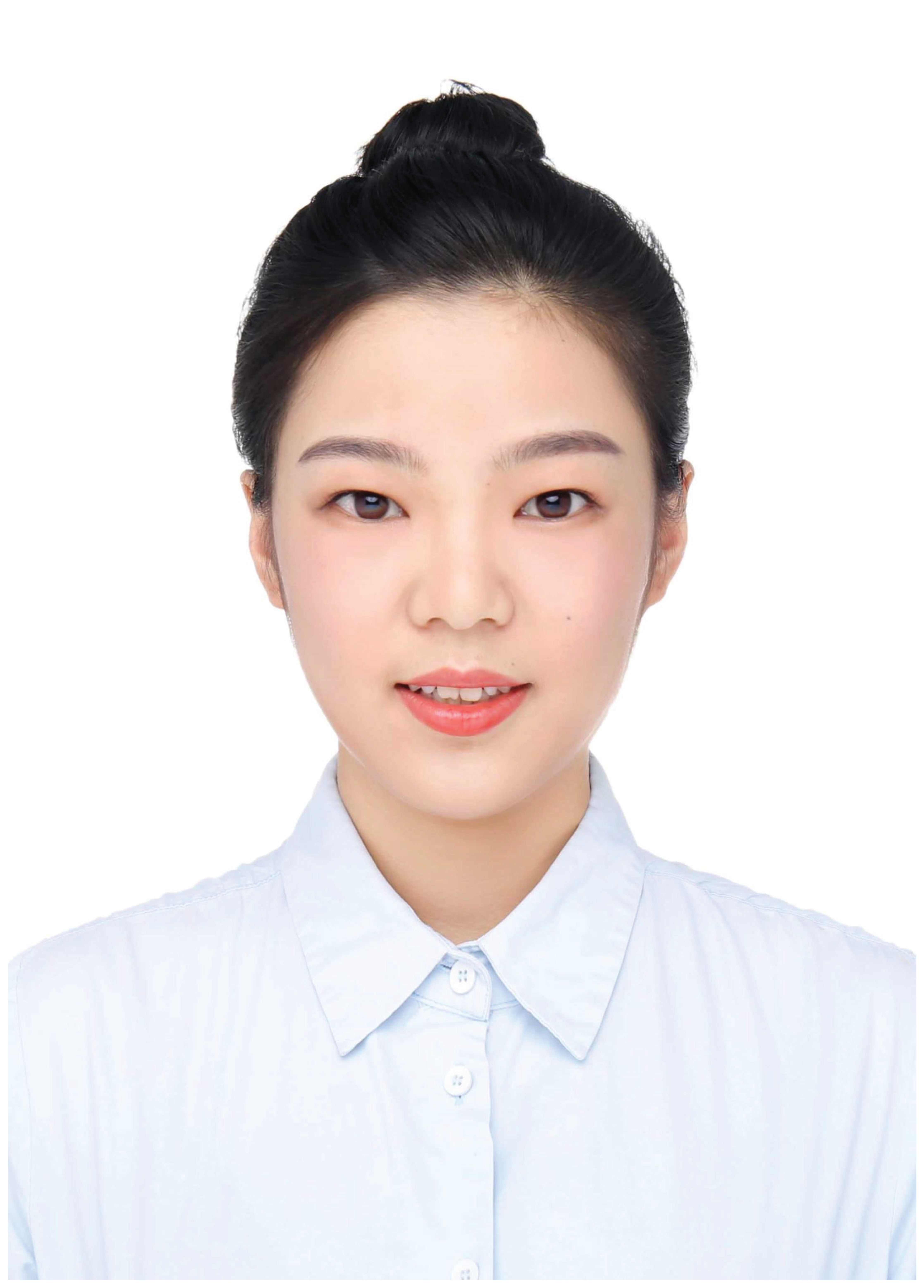}}]{Lili Liang}
	received her M.S. degree in Computer Science and Technology from Harbin University of Science and Technology in 2019. She is currently pursuing a Ph.D. in Computer Application Technology at the same institution. Her research interests include machine learning and video question answering (VideoQA).
\end{IEEEbiography}
\vfill

\end{document}